\pdfoutput=1

\documentclass[11pt]{article}

\usepackage[preprint]{acl}
\usepackage{amsmath}
\usepackage{amssymb}
\usepackage{amsthm}

\usepackage{times}
\usepackage{latexsym}
\usepackage{tabularx}
\usepackage{booktabs}
\usepackage{multirow}
\DeclareMathOperator*{\argmin}{arg\,min}

\usepackage{mathtools}

\usepackage{enumitem}

\usepackage{setspace}
\usepackage[T1]{fontenc}

\usepackage{colortbl} 
\usepackage{amssymb}
\usepackage[utf8]{inputenc}

\usepackage{microtype}

\usepackage{inconsolata}

\usepackage{graphicx}
\usepackage{hyperref}

\newcommand{\fyq}[1]{\textcolor{black}{#1}}

\title{\fyq{MergeIT: From Selection to Merging for Efficient Instruction Tuning}  }

\author{
 \textbf{Hongyi Cai\textsuperscript{1,2}\thanks{This work is done during internship at Shanghai Jiao Tong University.}},
  \textbf{Yuqian Fu\textsuperscript{3}},
  \textbf{Hongming Fu\textsuperscript{1}},
 \textbf{Bo Zhao\textsuperscript{1}\thanks{Corresponding Author: Bo Zhao}}
\\
 \textsuperscript{1}Shanghai Jiao Tong University, \;
 \textsuperscript{2}Universiti Malaya \\
 \textsuperscript{3}INSAIT, Sofia University "St. Kliment Ohridski"\\
 \small{{s2175463@siswa.um.edu.my}, {bo.zhao@sjtu.edu.cn}}
}

\begin{document}
\maketitle
\begin{abstract}
\fyq{Instruction tuning is crucial for optimizing Large Language Models (LLMs), yet mainstream data selection methods heavily rely on LLMs as instruction quality scorers, leading to high computational costs and reduced data diversity. To address these limitations, we propose \textbf{MergeIT}, a novel LLM-based \textbf{Merging} strategy for better \textbf{I}nstruction \textbf{T}uning that shifts the focus from selection to synthesis.
MergeIT operates in two stages: first, topic-aware filtering clusters and refines the dataset, preserving diversity while eliminating redundancy without relying on LLM-based scoring. Second, LLM-based merging synthesizes semantically similar instructions into more informative and compact training data, enhancing data richness while further reducing dataset size.
Experimental results demonstrate that MergeIT enables efficient, diverse, and scalable instruction selection and synthesis, establishing LLM-based merging as a promising alternative to conventional scoring-based selection methods for instruction tuning. Our source code and datasets are now available at \href{https://github.com/XcloudFance/MergeIT}{https://github.com/XcloudFance/MergeIT}}
\end{abstract}

\section{Introduction}
\fyq{Instruction tuning has emerged as a key technique for enhancing the adaptability and performance of Large Language Models (LLMs) \cite{dubey2024llama, jiang2023mistral}. By fine-tuning LLMs on carefully curated instruction datasets, e.g. \texttt{Alpaca\_52k} \cite{alpaca}, a dataset of 52,000 instructions and demonstrations generated by \texttt{text-davinci-003} \cite{gpt}., models can generalize across diverse tasks and prompts. However, selecting high-quality instruction data remains a critical challenge, as the choice of data directly affects fine-tuning outcomes.
}

\fyq{Existing selection methods \cite{chen2023alpagasus} primarily rely on LLMs as instruction quality \textbf{scorers}, ranking and filtering instructions based on predefined metrics \cite{deita,li2024superfiltering}. While this approach retains high-quality instructions, it introduces two major limitations. First, LLM-based scoring is computationally expensive, making large-scale selection infeasible. Second, scoring-based methods often prioritize high-ranked instructions at the expense of diversity, leading to redundant datasets that lack broader generalization. These trade-offs limit the scalability and effectiveness of instruction tuning.
}

\begin{figure}[t]
  \includegraphics[width=1.0\linewidth]{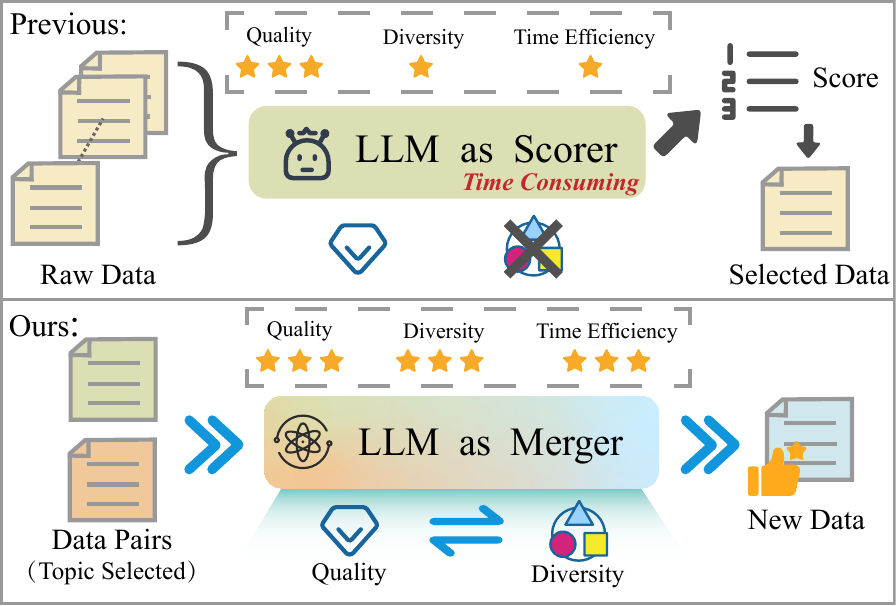} \hfill
  \caption{
  Comparison between our method and prior works. 
  \fyq{Unlike prior works that primarily use LLMs as scorers, we novelly explore their role as mergers, enhancing diversity and time efficiency.}
  }
  \vspace{-5mm}
      \label{fig:intro}
\end{figure}

\fyq{To overcome these challenges, we introduce \textbf{MergeIT}, \textit{a novel LLM-based merging strategy for better instruction tuning that moves beyond selection and leverages LLMs as \textbf{synthesizers} rather than mere \textbf{scorers}.} As illustrated in Fig.~\ref{fig:intro}, instead of scoring every single instruction, MergeIT merges semantically related instructions to create more informative, compact, and diverse samples. Comapred to the prior selection-based approaches, MergeIT improves dataset diversity while reducing time cost.
}

\fyq{Particularly, Our approach consists of two core stages: 1) Topic-aware Instruction Filtering – Instead of relying on LLM-based scoring, we first cluster the dataset into semantically meaningful topics and remove redundant instructions within each topic, ensuring diversity while preserving informativeness. 2) LLM-based Instruction Merging – Rather than eliminating similar instructions, we use LLMs to synthesize richer instructions by merging them into a more expressive and information-dense version. This step enhances dataset quality while reducing its size, making fine-tuning both more efficient and effective. 
By integrating topic-aware filtering with LLM-based merging, MergeIT achieves a faster, stronger, and more diverse instruction selection process, improving both the quality and efficiency of instruction tuning.}

\fyq{Extensive experiments are conducted to validate the effectiveness of MergeIT. Our method achieves state-of-the-art (SOTA) performance across six datasets, demonstrating significant improvements over existing approaches. Notably, our results highlight the feasibility and advantages of leveraging LLMs for instruction merging, a novel direction beyond traditional scoring-based selection. Our main contributions can be summarized as follows,}

\fyq{
\begin{itemize}
    \item 
    We propose \textbf{MergeIT}, a novel instruction data optimization framework that shifts from \textit{selection to synthesis}, integrating topic-aware filtering and LLM-based merging to enhance instruction tuning.
    \item 
   For the first time, we explore the use of LLMs to generate new instructions by merging two similar ones, offering novel insights for the instruction selection and generation.
    \item 
   Extensive experiments demonstrate the efficacy of our method, achieving high-quality, diverse instruction selection with reduced computational cost.
\end{itemize}
}

\section{Methodology}
\noindent\textbf{Problem Setup:} Given an initial instruction dataset $\mathcal{D} = \{(x_i, y_i)\}_{i=1}^n$ from \texttt{Alpaca\_52k}, where $x_i$ represents the input instruction and $y_i$ denotes the corresponding output, our goal is to select a high-quality subset $\mathcal{D}' \subseteq \mathcal{D}$ for instruction tuning, 
\fyq{such that $|\mathcal{D}'| \ll |\mathcal{D}|$.}

\subsection{Overview}\label{sec:overview}
\fyq{To ensure both diversity and quality in the selected subset while maintaining an efficient selection process, we propose \textbf{MergeIT} (\textbf{LLM-based Merging Strategy for Better Instruction Tuning}). The overview of our MergeIT method is illustrated in Fig.~\ref{fig:pipeline}. Our approach consists of two main steps: 1) Topic-aware Filtering; 2) LLM-based Merging.}

\fyq{Given an initial dataset $\mathcal{D}$, the first step filters out redundant examples by selecting only the most informative ones within each topic. This topic-aware strategy naturally ensures that the remaining samples $\mathcal{D}'$ are semantically diverse. Furthermore, by avoiding the use of large LLMs in this stage, we achieve an efficient filtering process. To further compress the data, the second step leverages LLMs to merge similar instances. By harnessing the strong comprehension and generation capabilities of LLMs, we synthesize high-quality and information-rich merged instructions. As a result, the size of $\mathcal{D}'$ is approximately halved, forming the final subset $\hat{\mathcal{D}}$. Unlike prior methods that utilize LLMs solely as scorers , we are the first to explore their potential in a merging framework.}

\subsection{Topic-aware Filtering}
\fyq{As stated in Sec.\ref{sec:overview}, our first step aims to remove redundant examples while preserving the diverse structural semantics of the initial dataset $\mathcal{D}$. To achieve this, our topic-aware filtering first clusters the instructions into specific topics (detailed in Sec.\ref{K-means}) and then selects the most representative subset from each topic, as described in Sec.~\ref{facility}.}




\subsubsection{K-means Clustering}
\label{K-means}
 We construct the initial clusters of data by introducing K-means algorithm to group pairs according to their instructions. All sentences are represented in embeddings space calculated from \texttt{all-MiniLM-L6-v2}, mapping into 384 dimensional features to calculate distances between samples. Given the instruction dataset $\mathcal{D} = \{(x_i, y_i)\}_{i=1}^n$, we first represent each instruction-output pair as a feature vector $f(x,y) \in \mathbb{R}^d$. To partition $\mathcal{D}$ into $m$ topically coherent clusters, we optimize:

\begin{align}
   \min_{\{\mathcal{D}_j\}_{j=1}^m} &\sum_{j=1}^m \sum_{(x,y) \in \mathcal{D}_j} \|f(x,y) - \mu_j\|_2^2 \nonumber \\
   \text{s.t.} \quad &\bigcup_{j=1}^m \mathcal{D}_j = \mathcal{D} \nonumber \\
   &\mathcal{D}_i \cap \mathcal{D}_j = \emptyset, \quad \forall i \neq j
\end{align}

where $\mu_j = \frac{1}{|\mathcal{D}_j|}\sum_{(x,y) \in \mathcal{D}_j} f(x,y)$ is the centroid of topic $t_j$.

\begin{equation}
   t(x,y) = \argmin_{j} \|f(x,y) - \mu_j\|_2^2
\end{equation}

\begin{figure}[t]
  \includegraphics[width=.95\linewidth]{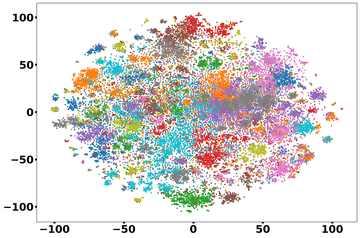} \hfill
  \caption {
  t-SNE visualization of K-means on \fyq{\texttt{Alpaca\_52k}. The instructions naturally form distinct clusters, indicating an inherent topical structure effectively captured by clustering.}
  }
    \vspace{-2mm}
      \label{fig:K-means}
\end{figure}

This provides the topic partitioning $\mathcal{T} = \{t_1, t_2, ..., t_m\}$ that maximizes intra-topic semantic coherence while ensuring clear boundaries between different instruction categories. To validate the feasibility of topic-based partitioning, we visualize the instruction embeddings using t-SNE \cite{vanDerMaaten2008} dimensionality reduction in Fig. \ref{fig:K-means}.

\subsubsection{Facility Location Function}
\label{facility}
To control the number of samples, we leverage facility location function as submodular function to select top $K$ representative data in each topic. For each topic cluster $\mathcal{D}_j$, the facility location function is defined as:

\begin{equation}
   F(\mathcal{D}'_j) = \sum_{(x,y) \in \mathcal{D}_j} \max_{(x',y') \in \mathcal{D}'_j} \text{sim}(f(x,y), f(x',y'))
\end{equation}

where $\mathcal{D}'_j \subseteq \mathcal{D}_j$ is the selected subset and $\text{sim}(\cdot, \cdot)$ measures the similarity between two instruction-output pairs in the feature space. This formulation aims to maximize:

\begin{align}
   \max_{\mathcal{D}'_j \subseteq \mathcal{D}_j} &\quad F(\mathcal{D}'_j) \nonumber \\
   \text{s.t.} &\quad |\mathcal{D}'_j| \leq K \nonumber \\
   &\quad Q(x,y) \geq q_j, \quad \forall (x,y) \in \mathcal{D}'_j
\end{align}

The facility location objective ensures each instruction in $\mathcal{D}_j$ is well-related their topics and remains representativeness in the selected subset $\mathcal{D}'_j$.

\begin{figure*}[t]
  \includegraphics[width=1.0\linewidth]{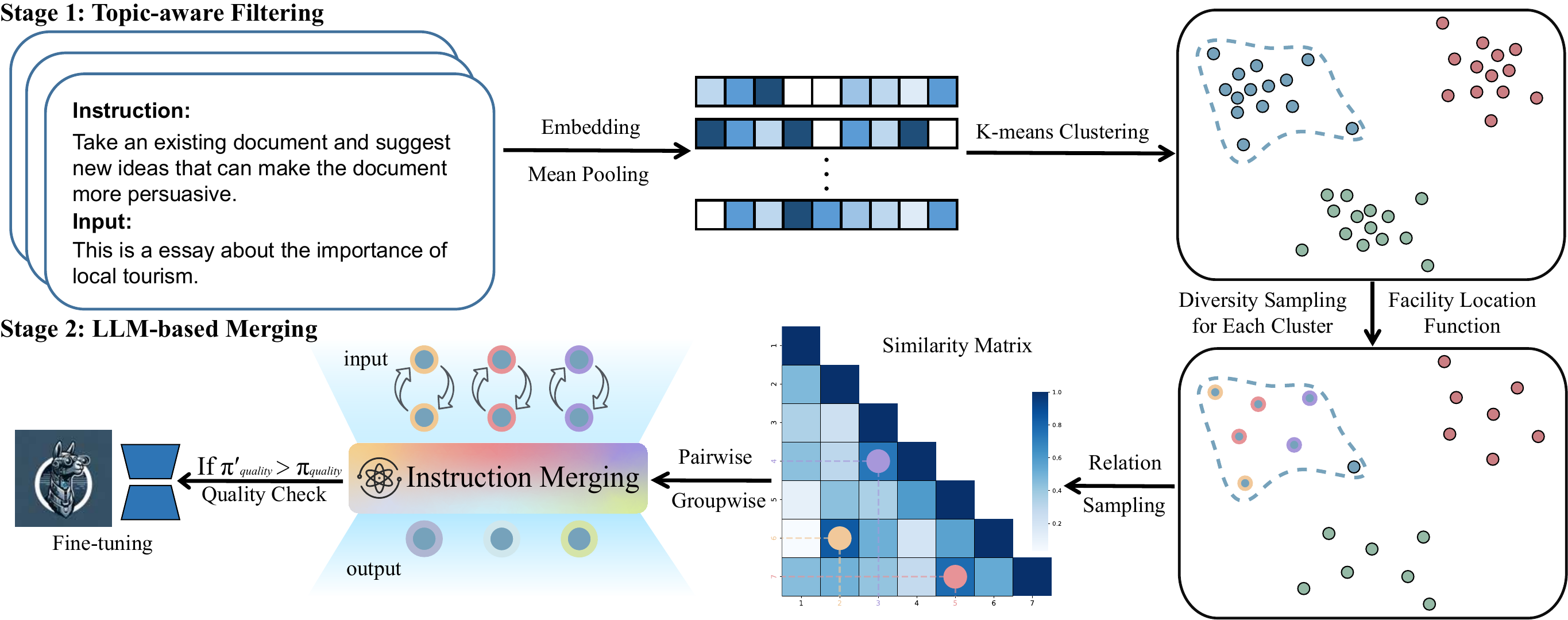} \hfill
  \caption {
  \fyq{Overview of MergeIT: 1) Topic-aware filtering clusters instructions into topics and filters redundant samples within each topic. 2) LLM-based merging synthesizes new instructions by combining similar pairs.}
  }
    \label{fig:pipeline}
\end{figure*}
\subsection{LLM-based Merging}
\label{MergeIT}
Following topic-based alignment and initial subset extraction $\mathcal{D}'_j$, the data volume reduces to approximately 20\% of the original corpus, effectively mitigating computational overhead for subsequent LLM processing. However, the resulting subset may still be less effective for LLM fine-tuning due to potential verbosity in remaining samples and quality inconsistencies arising from selection based solely on topical alignment without considering semantic relationships or response quality. 
\fyq{To address this, we for the first time propose the LLM-based merging,} a cluster-aware methodology that strategically combines semantically related instruction pairs from each topic $\{t_j\}_{j=1}^m$ through third-party LLMs. 


\fyq{As illustrated in Fig. \ref{fig:pipeline},}
starting with the 
$\mathcal{D}'_j$ \fyq{from the first filtering stage}, for each topic cluster $t_j$, we establish semantic equivalence classes through:

\begin{align}
  \mathcal{P}_j = \{&((x_i, y_i), (x_k, y_k)) \mid \nonumber \\
  &\text{sim}(f(x_i), f(x_k)) \geq \tau, \nonumber \\
  &(x_i, y_i), (x_k, y_k) \in \mathcal{D}'_j\}
\end{align}

where $\mathcal{P}_j$ denotes the set of instruction pairs in topic $t_j$ that exceed the similarity threshold $\tau$, $f(\cdot)$ denotes instruction embedding projection and $\text{sim}(\cdot,\cdot)$ computes cosine similarity. The merging operator $M:\mathcal{D}'_j \times \mathcal{D}'_j \rightarrow \hat{\mathcal{D}}$ employs an LLM-based synthesizer:

\begin{align}
   &M((x_i, y_i), (x_k, y_k)) = \text{LLM}_{\text{merge}}(\mathbf{c}_{ik}) = (\hat{x}, \hat{y})
\end{align}

where $\mathbf{c}_{ik} = [x_i; y_i; x_k; y_k]$ denotes the concatenated instruction-output context.

The final training protocol comprises:

\begin{align}
   \hat{\mathcal{D}} &= \bigcup_{j=1}^m \{M(p) \mid p \in \mathcal{P}_j\} \\
   \mathcal{L}(\theta) &= \mathbb{E}_{(\hat{x}, \hat{y}) \sim \hat{\mathcal{D}}} [-\log p_{\theta}(\hat{y}|\hat{x})]
\end{align}

where $\hat{\mathcal{D}}$ represents the merged instruction set from all topics and $\mathcal{L}(\theta)$ defines the fine-tuning objective for model parameters $\theta$.

\textbf{Quality Checking.} To prevent possible degradation during the merging process, we propose a quality preservation constraint:

\begin{equation}
    \pi'_{quality}(M_{i,j}) > \alpha(\pi_{quality}(S_i) + \pi_{quality}(S_j))
\end{equation}

\noindent where $M_{i,j}$ represents the merged result of samples $S_i$ and $S_j$, $\pi'_{quality}$ denotes the quality score after merging, $\pi_{quality}$ represents the quality score before merging, and $\alpha \in (0,1)$ is a parameter controlling the quality threshold, which we set by default at 0.75. This efficient quality checking mechanism ensures that merging operations only proceed when the resultant quality surpasses the weighted quality of the original samples. Our empirical evaluation demonstrates that this quality assessment process incurs negligible time consuming (0.5-1.0 seconds per sample pair), making it particularly suitable for integration into the main merging pipeline without introducing significant processing delays.

This cluster-constrained merging owns two computational advantages. First, it eliminates the $O(n^2)$ pairwise similarity bottleneck by restricting comparisons within pre-clustered topics ($|\mathcal{D}'_j| \ll |\mathcal{D}|$). The localized processing enables: (1) efficient identification of task-specific instruction patterns through intra-cluster analysis, and (2) context-aware merging of logically compatible samples – for instance, consolidating parallelizable tasks into batched instructions or combining complementary reasoning steps into coherent workflows. Second, the synthesis process inherently compresses lexical redundancies through LLM-based paraphrasing, simultaneously enhancing dataset density and instruction quality – a critical improvement over naive concatenation approaches. Additional examples and analysis are provided in Sec. \ref{sec4}.

\begin{table*}[t!]
\centering
\small
\begin{tabular}{|p{0.47\textwidth}|p{0.47\textwidth}|}
\hline
\multicolumn{1}{|c|}{\textbf{Pre-merge Tasks}} & \multicolumn{1}{c|}{\textbf{Post-merge Tasks}} \\
\hline
\multicolumn{2}{|c|}{\textit{Translation Task Group}} \\
\hline
\textbf{Task 1:} & \multirow{6}{0.47\textwidth}{\textbf{Merged Instruction:} "Translate the given phrase either from Spanish to English or English to Spanish" \\ 
\textbf{Input:} "The sun is not shining today. / La vida es un viaje, no un destino." \\
\textbf{Output:} "El sol no brilla hoy. / Life is a journey, not a destination." \\
\textbf{Quality Score:} 2.57} \\
Instruction: Translate to Spanish & \\
Input: "Life is a journey, not a destination." & \\
Output: "La vida es un viaje, no un destino." & \\
Quality Score: 1.88 & \\
\cline{1-1}
\textbf{Task 2:} & \\
Instruction: Translate to English & \\
Input: "El sol no brilla hoy." & \\
Output: "The sun is not shining today." & \\
Quality Score: 1.93 & \\
\hline
\multicolumn{2}{|c|}{\textit{Language Analysis Task Group}} \\
\hline
\textbf{Task 1:} & \multirow{6}{0.47\textwidth}{\textbf{Merged Instruction:} "Identify the type of sentence structure used and analyze its phrase for the part of speech" \\
\textbf{Input:} "The birds are singing, but the sun is hidden on this windy day" \\
\textbf{Output:} "The given sentence '...' is a Compound-complex sentence. Also, the phrase 'windy day' within this sentence is an adjective phrase, as it describes the noun 'day' indicating that it is windy." \\
\textbf{Quality Score:} 3.13} \\
Instruction: Categorize part of speech & \\
Input: "very" & \\
Output: "The word 'very' is an adverb." & \\
Quality Score: 1.71 & \\
\cline{1-1}
\textbf{Task 2:} & \\
Instruction: Identify sentence structure & \\
Input: "The birds are singing, but the sun is hidden." & \\
Output: "Compound-complex sentence" & \\
Quality Score: 1.23 & \\
\hline
\multicolumn{2}{|c|}{\textit{Calculation Task Group}} \\
\hline
\textbf{Task 1:} & \multirow{6}{0.47\textwidth}{\textbf{Merged Instruction:} "Determine if the given number is odd or even, and also check if it is a prime number." \\
\textbf{Input:} "Number: 48" \\
\textbf{Output:} "The number 48 is even. It is not a prime number because it has divisors other than 1 and itself." \\
\textbf{Quality Score:} 3.08} \\
Instruction: Identify if the given number is odd or even. & \\
Input: "Number: 48" & \\
Output: "Even" & \\
Quality Score: 2.35 & \\
\cline{1-1}
\textbf{Task 2:} & \\
Instruction: Is the given number a prime number? & \\
Input: "19" & \\
Output: "Yes, 19 is a prime number." & \\
Quality Score: 2.97 & \\
\hline
\end{tabular}
\caption{Comparison of Pre-merge and Post-merge Language Tasks}

  \vspace{-3mm}
\label{tab:merge-comparison}
\end{table*}


\section{Does Merging Really Make the Data Better for Training?}
\label{sec4}
In this section, we further delve into the real examples of merging in \texttt{Alpaca\_52k}, to demonstrate the post-processed outcomes, as well as providing analysis to its practical feasibility.
As mentioned before, we introduce \texttt{Deita} scorer \cite{deita} as the measurement of quality of training data from the instruction and its response. To clearly reveal the examples of language tasks, we 
picked three instruction groups (e.g., Translation, Language Analysis, and Calculation), shown in Tab.~\ref{tab:merge-comparison}.

\fyq{To further validate our approach of instruction merging, we explore three key questions:}
1) Does instruction merge improve the overall quality? 2) How does 
the merged data
impact the explanation depth of responses? 3) Does 
LLM
effectively integrate knowledge from different components?

\textbf{Quality Enhancement.} Our empirical analysis reveals a consistent pattern of quality improvement across all examined task groups. As shown in Tab.~\ref{tab:merge-comparison}, the quality scores of merged instructions consistently surpass 
\fyq{the initial ones.}
Specifically, in the \textit{Translation Task Group}, we observe an increase from an average score of 1.90 to 2.57, representing a 35.2\% improvement. Similar enhancements are evident in \textit{Language Analysis} (141.7\% increase) and \textit{Calculation} (15.8\% increase) tasks. These substantial improvements in quality scores suggest that our merge strategy effectively combines individual instructions while maintaining coherence.

\textbf{Response Depth.} A notable outcome of our merging process is the improvement in response depth and completeness. Post-merge responses demonstrate a marked increase in explanatory content and reasoning clarity. 

For instance, in \textit{Language Analysis Task Group}, since pre-merge instructions are logically coherent with each other, like "categorize part of speech" and "identify sentence structure" all related to the analytical of sentences, LLMs not only connect instructions by simply adding "and" in between, but also emerge to provide more comprehensive outputs. In addition, the Task 1 in the Group is promoted to perform analysis for the whole sentences, rather than merely focusing on a single word, which provides more insights to increment the quality from its original plain requests.

\textbf{Knowledge Integration.} Our analysis further demonstrates the effectiveness of knowledge integration across different task components. The merge process successfully preserves the essential elements of individual tasks while creating cohesive instructions. This is particularly evident in the \textit{Calculation Task Group}, where the merged instruction seamlessly combines number property analysis (odd/even) with prime number verification. The resulting quality score of 3.08 and comprehensive response ("The number 48 is even. It is not a prime number because it has divisors other than 1 and itself") validates that this integration not only maintains but enhances the overall task effectiveness.

\section{Experiments}

In this section, we evaluate two open-source and commonly used models: \texttt{Mistral-7b-v0.3} \cite{jiang2023mistral} and \texttt{LLaMA3-8b} \cite{dubey2024llama} with two types of benchmarks: LLM-based (MT-Bench \cite{mtbench}, AlpacaEval \cite{instructeval}), and Huggingface Open LLM Leaderboard (Hellaswag \cite{hellaswag}, MMLU \cite{mmlu}, GSM8k \cite{gsm8k}, ARC \cite{arc}, and TruthfulQA \cite{truthfulqa}). 
\fyq{Several baseline methods, including Alpaca-52k (full dataset), Superfiltering \cite{li2024superfiltering}, Random selection, Perplexity, K-means, and LIMA, are included for comparison.}

\begin{table*}[t]
    \centering
    \renewcommand{\arraystretch}{1.1}
    \setlength{\tabcolsep}{5pt}
    \begin{tabular}{l|c|cccccc}
        \hline
        {Model} & MT-Bench & \multicolumn{5}{c}{Huggingface Open LLM Leaderboard (Acc.) $\uparrow$}   \\
        & Score & Hellaswag & MMLU & GSM8k & ARC & TruthfulQA & Average \\
        \hline
        Mistral-7b-v0.3 & 3.639 & 60.94 & 58.96 & 36.62 & 48.81 & 22.44 & 45.55 \\
        \hline
        Alpaca-52k & 4.018 & 61.18 & 57.73 & 31.61 & 53.07 & 28.76 & 46.47 \\
        SuperFiltering-10\% & 3.963 & 60.98 & 59.34 & 35.71 & 49.83 & 29.99 & 47.17 \\
        Random-6k & 4.314 & 60.83 & 58.75 & 35.03 & 53.07 & 32.19 & 47.97 \\
        Perplexity-6k & 4.352 & \textbf{61.64} & 58.48 & \underline{37.00} & 51.88 & 31.21 & \underline{48.04} \\
        K-means-6k & 4.283 & 60.86 & 58.45 & 35.10 & 52.05 & 32.46 & 47.78 \\
        LIMA-6k & \underline{4.440} & 60.58 & \textbf{59.34} & 34.34 & \underline{53.33} & 31.82 & 47.88 \\
        \hline
        \rowcolor{gray!30} \textbf{MergeIT-6k (Ours)} & \textbf{4.481} & \underline{61.40} & \underline{59.01} & \textbf{37.30} & \textbf{54.95} & \textbf{33.41} & \textbf{49.21} \\
        \hline
        LLaMA3-8b & 3.418 & 60.17 & \textbf{62.13} & \underline{50.42} & 50.26 & 26.93 & 49.98 \\
        \hline  
        Alpaca-52k & 3.718 & 60.57 & 61.36 & 46.10 & \underline{53.41} & 30.72 & 50.43 \\
        SuperFiltering-10\% & 3.968 & 60.38 & \textbf{61.95} & 50.34 & 51.54 & 29.87 & 50.82 \\
        Random-6k & 3.912 & 60.83 & 58.75 & 35.03 & 53.07 & 32.44 & 48.02 \\
        Perplexity-6k & 4.120 & \underline{61.14} & 61.09 & 50.87 & 53.50 & 31.33 & \underline{51.58} \\
        K-means-6k & 3.731 & 60.86 & 58.45 & 35.10 & 53.07 & 32.31 & 47.96 \\
        LIMA-6k & \underline{4.450} & 60.58 & 61.28 & 50.34 & 51.11 & \underline{34.27} & 51.51 \\
       \rowcolor{gray!30}\textbf{MergeIT-6k (Ours)} & \textbf{4.525} & \textbf{61.96} & \underline{61.49} & \textbf{51.87} & \textbf{54.95} & \textbf{34.52} & \textbf{52.96} \\
        \hline
    \end{tabular}
    \caption{Performance comparison on standard benchmarks (Experiment A). The best results are highlighted in \textbf{bold}, and the second-best results are \underline{underlined}.}
    \label{experiment}
\end{table*}

\subsection{Experimental Setup}
We adopt LLaMA-Factory\footnote{https://github.com/hiyouga/LLaMA-Factory}  as our training base. All fine-tuning experiments utilize LoRA \cite{hu2021lora} with the learning rate $2 \times 10^{-5}$, 3 epochs, a batch size of 4, and Cosine scheduler with warmup ratio 0.5. Additionally, we apply 4 pieces of Huawei Ascend 910b 64GB, to train the models. To evaluate Open LLM Leaderboard, we use lm-evaluation-harness\footnote{https://github.com/EleutherAI/lm-evaluation-harness} as it integrates all of the required benchmarks.

\textbf{Baselines Setup.} For selected baseline models, we extensively experimented with some traditional data selection paradigms. Specifically, K-means baseline clustered the data into 120 groups and selected the most distant samples from each centroid of clusters to realize diversity manner. Perplexity-based method calculates the perplexity score from LLaMa3-8b, formulated below: 
\begin{equation}
    PP(W) = 2^{-\frac{1}{N}\sum_{i=1}^{N}\log_2P(w_i|w_1,...,w_{i-1})} ,
\end{equation}
where $PP(W)$ represents the perplexity score for sequence $W$, $N$ is the sequence length, and $P(w_i|w_1,...,w_{i-1})$ denotes the conditional probability of predicting the current word $w_i$ given all previous words. For SuperFiltering, we reuse their 10\% filtered data from their Instruction-Following Difficulty score.

\subsection{Data Scaling}
\begin{figure}[h!]
  \includegraphics[width=1.0\linewidth]{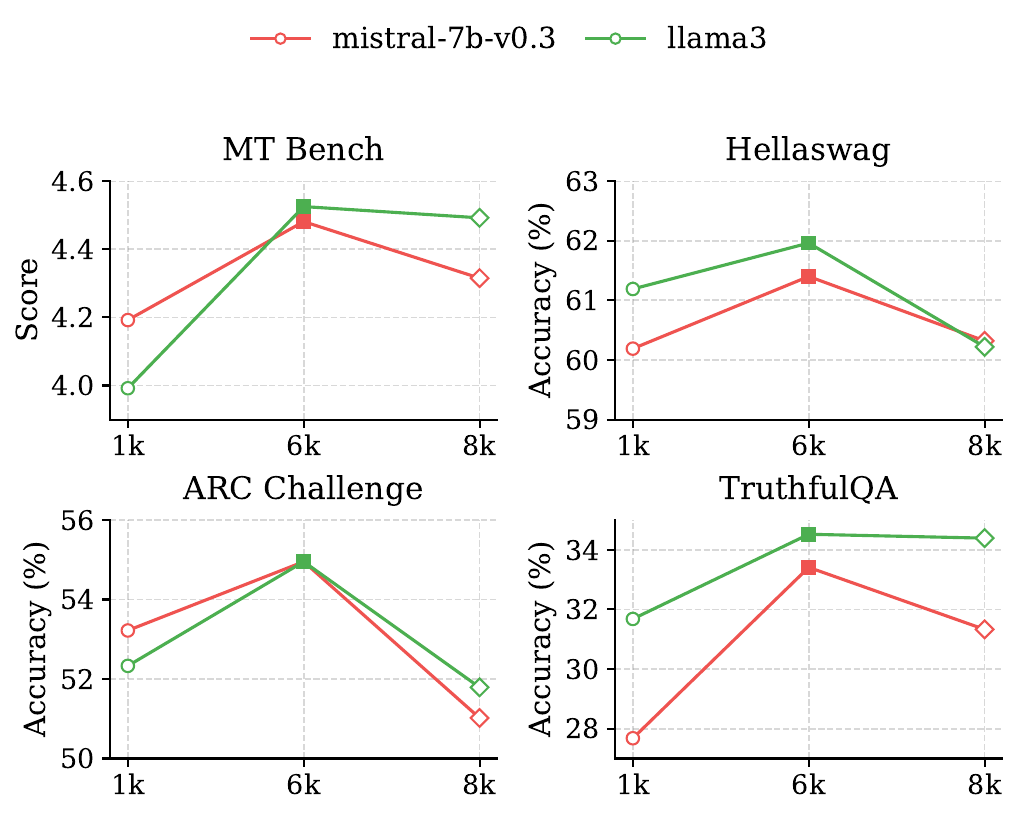} \hfill
  \caption {The figure shows the comparison between different scales of number of data in instruction tuning. }
  \vspace{-5mm}
      \label{fig:scaling}
\end{figure}

To validate the most applicable number of instruction tuning samples, we implement experiments on testing different data scaling. We include approximately 1k, 6k and 8k of merged data as training subsets and also evaluate on  MT-Bench, Hellaswag, ARC and TruthfulQA, as illustrated in Fig. \ref{fig:scaling}. In the given graph, we finalize our corpus size into 6k since it reveals the best trade-off between performance and efficiency, while 9k however demonstrates reverse effects even if the data scales up and 1k data is likely causing the lack of robustness and understanding. 

\subsection{Main Results}
\begin{figure}[h]
  \includegraphics[width=1.0\linewidth]{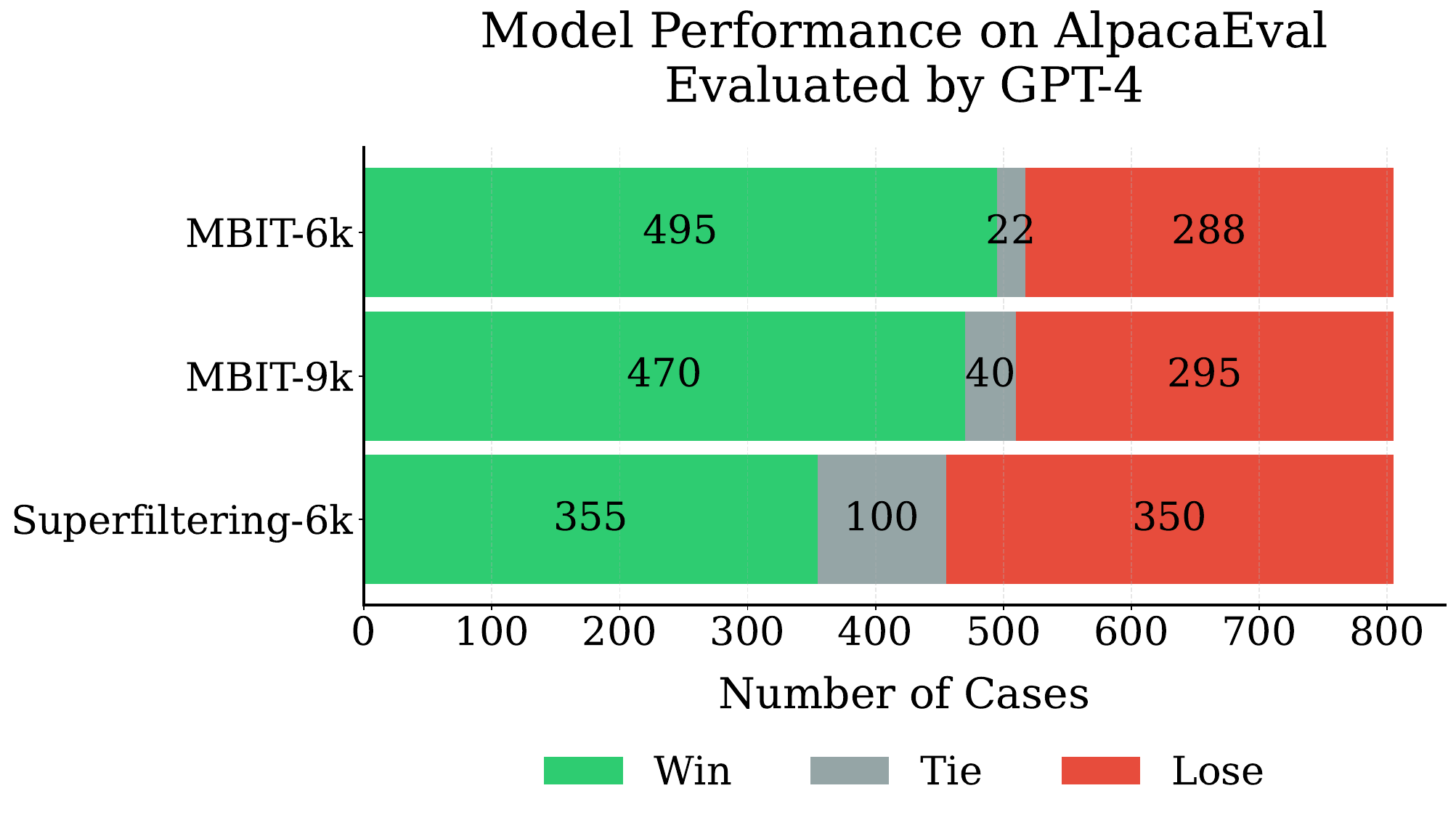} \hfill
\vspace{-5mm}

  \caption {AlpacaEval results. Compared models are MergeIT-6k V.S Alapca-52k full samples (line 1), MergeIT-9k V.S Alapca-52k full samples (line 2) and Superfiltering-6k V.S Alapca-52k full samples (line 3)}
    \vspace{-3mm}
      \label{fig:alpacaeval}
\end{figure}

\textbf{LLM as a Judge.}
As shown in Table \ref{experiment}, MergeIT-6k achieves the highest performance in LLM scoring, reaching 4.481 and outperforming other baselines by up to 0.518 on Mistral-7b-v0.3 (a 0.842 improvement from base model), while achieving 4.525 on LLaMA3-8b (a 1.107 improvement from base model), surpassing other methods by up to 0.807. Further evaluation on AlpacaEval \cite{alpaca_eval} in Fig. \ref{fig:alpacaeval} using GPT-4 shows MergeIT-6k winning in 485 out of 800 comparisons against SuperFiltering-6k's 355 wins, confirming its effectiveness across different judges.

\textbf{Huggingface Open LLM Leaderboard.} Our MergeIT achieves state-of-the-art performance (49.21\% average score) across all five tasks, outperforming strong baselines LIMA-6k (47.88\%) and K-means-6k (47.78\%). The improvements are particularly notable on ARC (54.95\%) and TruthfulQA (33.41\%). When trained on LLaMA-generated data, MergeIT further  
obtains 52.96\% average score, with significant gains on GSM8k (51.87\%) and ARC (54.95\%), demonstrating its effectiveness across diverse tasks.

\begin{table*}[htbp]
    \centering
    \scalebox{0.9}{  
    \renewcommand{\arraystretch}{1.1}
    \setlength{\tabcolsep}{5pt}
    \begin{tabular}{ccc|c|c|ccccc}
        \hline
         Merging & K-means & Quality & \# Samples & MT-Bench & Hellaswag & MMLU & GSM8k & ARC &  \\
        & & Checking & & Score & (Acc.) & (Acc.) & (Acc.) & (Acc.) & $\mu_{avg}$ \\
        \hline
          \checkmark & \checkmark & \checkmark & 6000  & 4.481 & 61.40 & 59.01 & 37.30 & 53.33 & 52.76 \\
        \hline
          & \checkmark & \checkmark & 12000 & 4.300 & 60.31 & 59.06 & 34.42 & 51.37 & 51.29 \\
         \checkmark & & \checkmark & 6000 & 4.112 & 59.68 & 58.75 & 30.29 & 50.01 & 49.68 \\
         \checkmark & \checkmark & & 6000 & 4.081 & 59.81 & 58.76 & 33.16 & 49.56 & 50.32 \\
        \hline
    \end{tabular}
    }  

    \caption{Ablation studies on different components of our method.}
    \label{tab:ablation}

\end{table*}

\begin{table*}[htbp]
    \centering
    \renewcommand{\arraystretch}{1.1}
    \setlength{\tabcolsep}{5pt}
    \begin{tabular}{l|c|ccccc}
        \hline
         Merging Methods & MT-Bench & Hellaswag & MMLU & GSM8k & ARC &  \\
        & Score & (Acc.) & (Acc.) & (Acc.) & (Acc.) & $\mu_{avg}$ \\
        \hline
         MergeIT Merging - Mistral-7b & 4.481 & 61.40 & 59.01 & 37.30 & 53.33 & 52.76 \\
         \hline
        Random Select within Topics & 4.198 & 59.98 & 59.05 & 35.25 & 50.68 & 51.24 \\
        \hline
         Concat Merging & 4.200 & 60.16 & 58.76  & 36.47 & 50.85 & 51.56 \\
        \hline
    \end{tabular}
    \caption{Ablation studies on different merging methods.}
    \label{tab:ablation2}
\end{table*}

\subsection{Ablation Study}
To better understand the contribution of each component in our method, we conduct comprehensive ablation studies as shown in Tab.~\ref{tab:ablation} and Tab. \ref{tab:ablation2}. Our full model with all three components (6000 samples) achieves the best overall performance with an average accuracy of 52.76\% and MT-Bench score of 4.481.

\textbf{Merging Samples from Given Pairs.} We first investigate the impact of our merging strategy. When removing the merging component (12000 samples), the average accuracy drops by 0.84\% and MT-Bench score decreases to 4.300, suggesting that our merging strategy effectively enhances model performance by providing more diverse training samples.

\textbf{Diversity in Topics.} The K-means clustering plays a crucial role in maintaining diversity. Without K-means but keeping quality checking (second row), the model's performance drops significantly across all metrics, particularly on GSM8k (-7.01\%) and ARC (-3.32\%). This indicates that K-means clustering helps ensure a balanced representation of different topics in the training data.

\textbf{Quality Assurance.} 
\fyq{Removing the quality-checking component (last row) results in the most significant performance drop, with the MT-Bench score decreasing to 4.081 and average accuracy declining by 2.44\%. The impact is particularly pronounced in reasoning-heavy tasks such as ARC (-3.77\%) and GSM8K (-4.14\%), highlighting the critical role of quality checking in maintaining high-quality training samples and ensuring robust model performance.}

\textbf{Different Merging Strategies.} 
\fyq{We further compare MergeIT's merging with two alternative strategies in Tab.~\ref{tab:ablation2}: random selection within topics and simple concatenation-based merging. While random selection preserves topic awareness, it lacks merging, resulting in a performance drop (51.24\% average accuracy). Simple concatenation using "and" as the connector performs slightly better (51.56\%) but still lags behind MergeIT's merging (52.76\%), validating our idea of using LLM-guided merging for generating high-quality samples.}

\section{Related Work}

\subsection{Instruction Data Selection}
With the rapid development of large language models in recent years and the thorough exploration of large-scale training data, research has shifted from model-centric to data-centric approaches. For instance, \citealp{zhou2024lima} has demonstrated that efficient alignment can be performed on small-scale high-quality data. Currently, the training data of most large language models (\citealp{dubey2024llama}) must be filtered in advance to ensure high quality, thereby improving both training effectiveness and efficiency. Regarding the critical task of selecting instruction fine-tuning data, in addition to traditional methods such as coreset selection (\citealp{zhang2024tagcos}) and clustering (\citealp{he2024efficient}), recent methods mainly propose new data quality indicators, struggling to balance data quality and diversity to achieve high-quality data selection.

\subsection{Quality Filtering}
The quality of instruction fine-tuning data—such as clarity of instructions and normativity of expressions—directly influences the final performance of a model. Common quality filtering methods are indicator-based, meaning each sample is assigned an index score, and high-scoring samples are selectively retained. Typical indicators include perplexity (\citealp{ankner2024perplexed, mekala2024smaller}), instruction-following difficulty (\citealp{li2023quantity, li2024superfiltering, li2024selective}), LLM-based scoring (\citealp{deita, song2024iterselecttune}), manual evaluation (\citealp{liu2024coachlm}), influence values (\citealp{xia2024less, liu2024less, yu2024mates}), and submodular functions (\citealp{agarwal2024delift, renduchintala2024smart}). Although these methods can effectively filter out relatively high-quality samples, they do not improve any shortcomings in the remaining samples or rely solely on LLM-based priors for enhancement, limiting the ultimate quality of the data, thus constraining model improvement. However, by merging instructions, we can fully leverage information from all samples so that they complement each other, thereby maximizing data quality and surpassing the original dataset.

\subsection{Diversity-Related Works}
Diversity in instruction fine-tuning data—covering domains, formats, and sources—is equally vital. Over-filtering for quality risks omitting multiple task types or domain knowledge, reducing the model’s generalization ability. Existing methods for instruction data selection often overlook diversity (\citealp{shen2024rethinking}) or rely on traditional methods like K-means sampling (\citealp{li2023quantity, ge2024clustering, maharana2024adapt}) and K-center greedy (\citealp{deita, wang2024your}), which often yield suboptimal results. Other diversity-focused strategies remain heuristic, such as the n-gram-based bidirectional graph used in \citealp{wu2024best}. However, these methods typically discard certain samples outright, inevitably reducing overall data diversity. In contrast, our instruction merging method retains information from all samples, effectively preserving diversity while maintaining quality.

\section{Conclusion}
\fyq{In this paper, we introduce MergeIT, a novel framework for efficient instruction tuning. By integrating topic-aware filtering and LLM-based merging, MergeIT effectively filters and combines instruction pairs, reducing dataset size while enhancing informational richness. Notably, we pioneer the use of LLMs for instruction merging, leveraging their generative capabilities to synthesize more informative and compact training data. Experimental results confirm the effectiveness of our merging strategy, achieving superior performance over all baselines and demonstrating the potential of LLMs beyond traditional scoring-based selection.}

\section{Limitation}
Our work remains several challenge to be resolved: 1) Our work remains inevitable clustering process to maintain the diversity in the context, which is still served as a less stable method when it comes to larger datasets. 2) Merging process occasionally loses the information given from the samples, which potentially harms the information from the original corpus.


\bibliography{custom}

\clearpage
\appendix

\section{Merging Examples}
We further cherry-pick some merging examples as well as corresponding pre-merge instructions to observe the effects brought by LLMs. More groups and scenarios are provided, shown in Tab. \ref{tab:merging2}

\section{Small-parameter Models for Merging}
\label{sec:appendix}

Even though MergeIT has saved the most budgets of invoking LLMs API vastly, it still incurs burdens from expensive API calling costs. To solve this challenge, we finetuned a small open-source model, Gemma2-9b from a small portion of data generated by GPT-4o examples. The evaluated results are shown in Tab. \ref{tab:mergingappend}. From the given table, Gemma2-9b holds same ability even if the size of model is rather smaller than GPT-4o, which proves the process is possible to be migrated to smaller LLMs.

\begin{table*}[htbp]
\centering
\small
\begin{tabular}{|p{0.47\textwidth}|p{0.47\textwidth}|}
\hline
\multicolumn{1}{|c|}{\textbf{Pre-fusion Tasks}} & \multicolumn{1}{c|}{\textbf{Post-fusion Tasks}} \\
\hline
\multicolumn{2}{|c|}{\textit{Customer Data Management Task Group}} \\
\hline
\textbf{Task 1:} & \multirow{6}{0.47\textwidth}{\textbf{Fused Instruction:} "Perform a customer-related task for a given company based on specific instructions." \\ 
\textbf{Input:} "Task: Update contact information\textbackslash nCompany: ABC Limited\textbackslash nCustomer: John Smith\textbackslash nNew Contact Number: 123-456-7890\textbackslash n\textbackslash nTask: Send a payment reminder email\textbackslash nCustomer: John Doe\textbackslash nAmount Owed: \$100" \\
\textbf{Output:} "1. The record for customer 'John Smith' has been updated in the customer database of ABC Limited by changing the contact number to 123-456-7890.\textbackslash n\textbackslash n2. Email to John Doe:\textbackslash n\textbackslash nDear John,\textbackslash n\textbackslash nWe hope this finds you well! We noticed that your payment of \$100 is past due..." \\
\textbf{Quality Score:} 2.89} \\
Instruction: Update a record in the customer database of a given company according to the instructions & \\
Input: "company: ABC Limited\textbackslash ninstructions: change the contact number for customer 'John Smith'" & \\
Output: "The record for customer 'John Smith' has been updated in the customer database of ABC Limited by changing the contact number." & \\
Quality Score: 2.31 & \\
\cline{1-1}
\textbf{Task 2:} & \\
Instruction: Write an email to remind a customer of their past due payment & \\
Input: "Name: John Doe, Amount owed: \$100" & \\
Output: "Dear John,\textbackslash n\textbackslash nWe hope this finds you well! We noticed that your payment of \$100 is past due..." & \\
Quality Score: 2.45 & \\
\hline
\multicolumn{2}{|c|}{\textit{Conflict Resolution Task Group}} \\
\hline
\textbf{Task 1:} & \multirow{6}{0.47\textwidth}{\textbf{Fused Instruction:} "Provide strategies for resolving conflicts in different settings" \\ 
\textbf{Input:} "Family disputes and angry customer emails" \\
\textbf{Output:} "To effectively resolve conflicts in both personal and professional settings, consider the following strategies: 1. Open Communication: Encourage all parties involved to express their thoughts..." \\
\textbf{Quality Score:} 2.95} \\
Instruction: Provide strategies for personal conflicts & \\
Input: "Family dispute situation" & \\
Output: "1. Open Communication: Encourage all parties involved to express their thoughts..." & \\
Quality Score: 2.43 & \\
\cline{1-1}
\textbf{Task 2:} & \\
Instruction: Handle professional conflicts & \\
Input: "Angry customer email" & \\
Output: "1. Acknowledge the issue 2. Express empathy..." & \\
Quality Score: 2.51 & \\
\hline
\multicolumn{2}{|c|}{\textit{Technical Support Task Group}} \\
\hline
\textbf{Task 1:} & \multirow{6}{0.47\textwidth}{\textbf{Fused Instruction:} "Provide troubleshooting guide for technical issues" \\ 
\textbf{Input:} "1. 'An unknown error occurred. Error code: 500.'\textbackslash n2. 'Your internet connection is slow.'" \\
\textbf{Output:} "To troubleshoot the error code 500, which is an internal server error, start by examining the server and application logs..." \\
\textbf{Quality Score:} 3.12} \\
Instruction: Detect entities from text & \\
Input: "Yesterday afternoon, Amazon Web Services went down..." & \\
Output: "Entities detected: Amazon Web Services, US-East-1 data center, Virginia." & \\
Quality Score: 2.34 & \\
\cline{1-1}
\textbf{Task 2:} & \\
Instruction: Generate tech support conversation & \\
Input: "Customer: I need to reset my password." & \\
Output: "Tech Support: No problem! What is the email address you use to login..." & \\
Quality Score: 2.41 & \\
\hline
\end{tabular}
\caption{Comparison of Pre-fusion and Post-fusion Language Tasks}
\label{tab:merging2}
\end{table*}

\begin{table*}[htbp]
    \centering
    \renewcommand{\arraystretch}{1.1}
    \setlength{\tabcolsep}{5pt}
    \begin{tabular}{l|c|ccccc}
        \hline
         Merging Methods & MT-Bench & Hellaswag & MMLU & GSM8k & ARC &  \\
        & Score & (Acc.) & (Acc.) & (Acc.) & (Acc.) & $\mu_{avg}$ \\
        \hline
         GPT-4o merging & 4.481 & 61.40 & 59.01 & 37.30 & 53.33 & 52.76 \\
        \hline
         Gemma2-9b merging & 4.370 & 59.05 & 60.98  & 36.25 & 50.68 & 51.74 \\
        \hline
        Concat Merging & 4.200 & 60.16 & 58.76  & 36.47 & 50.85 & 51.56 \\

        \hline
    \end{tabular}
    \caption{Merging methods comparisons. All generated subset of data are trained on Mistral-7b-v0.3.}
    \label{tab:mergingappend}
\end{table*}
\end{document}